\documentclass[sigconf]{acmart}
\renewcommand\footnotetextcopyrightpermission[1]{} 

\usepackage{subfigure}
\usepackage{multirow}
 \usepackage{booktabs}
 \usepackage{multirow}
\usepackage{xcolor} 
\definecolor{Grey}{rgb}{0.5,0.5,0.5}    
\usepackage{threeparttable}  

\usepackage{algorithm}
\usepackage{algorithmic}
\usepackage[capitalize]{cleveref}

\usepackage{graphicx}
\usepackage{amsmath}

\usepackage{amssymb}
\usepackage{booktabs}

\setcopyright{none}
\begin{document}

\title{Camera-Incremental Object Re-Identification with Identity Knowledge Evolution}

\author{Hantao Yao$^{1}$, Lu Yu$^{2}$, Jifei Luo$^{3}$, Changsheng Xu$^{1}$}
\affiliation{\institution{1 State Key Laboratory of Multimodal Artificial Intelligence Systems, Institute of Automation, CAS\\2 Tianjin University of Technology;3 University of Science and Technology of China}}
\email{hantao.yao@nlpr.ia.ac.cn}


\renewcommand{\algorithmicensure}{\textbf{Output:}}

\begin{abstract}
Object Re-identification (ReID) aims to retrieve the probe object from many gallery images with the ReID model inferred based on a stationary camera-free dataset by associating and collecting the identities across all camera views.
When deploying the ReID algorithm in real-world scenarios, the aspect of storage, privacy constraints, and dynamic changes of cameras would degrade its generalizability and applicability.
Treating each camera's data independently, we introduce a novel ReID task named Camera-Incremental Object Re-identification (CIOR) by continually optimizing the ReID mode from the incoming stream of the camera dataset.
Since the identities under different camera views might describe the same object, associating and distilling the knowledge of common identities would boost the discrimination and benefit from alleviating the catastrophic forgetting.
In this paper, we propose a novel Identity Knowledge Evolution (IKE) framework for CIOR, consisting of the Identity Knowledge Association (IKA), Identity Knowledge Distillation (IKD), and Identity Knowledge Update (IKU).
IKA is proposed to discover the common identities between the current identity and historical identities.
IKD has applied to distillate historical identity knowledge from common identities and quickly adapt the historical model to the current camera view.
After each camera has been trained, IKU is applied to continually expand the identity knowledge by combining the historical and current identity memories.
The evaluation of Market-CL and Veri-CL shows the Identity Knowledge Evolution (IKE) effectiveness for CIOR. 
code:https://github.com/htyao89/Camera-Incremental-Object-ReID
\end{abstract}




\maketitle

\section{Introduction}
\label{sec:intro}

Object Re-identification (ReID), such as person ReID~\cite{zheng2015scalable} and vehicle ReID~\cite{DBLP:conf/eccv/LiuLMM16}, aims to search the probe object from the many gallery images captured from different camera views, attracting increasing attention recently.
Most of the existing object ReID methods can be divided into supervised object ReID~\cite{Zhang_2020_CVPR, Chen_2021_ICCV, zhang2021person, xia2019second, chen2019mixed, quadruplet, large-margin-learning,he2019part,he2020multi,He_2021_ICCV,meng2020parsing} and unsupervised object ReID~\cite{Chen_2019_ICCV, Liu_2019_CVPR, zhong2018generalizing, fu2019self, Zhang_2019_ICCV, ge2020mutual,DBLP:conf/nips/Ge0C0L20,dai2021cluster,DBLP:conf/cvpr/ZhangG0021}, where the supervised object ReID aims to infer a discriminative ReID model with the annotated labels, and the unsupervised object ReID applies the pseudo-labels or unsupervised consistency constraint for representation learning. 
The existing object ReID assumes a camera-free stationary dataset with the images and identities all available during training, also called camera-free ReID.

\begin{figure}
  \centering
   \includegraphics[width=1.0\linewidth]{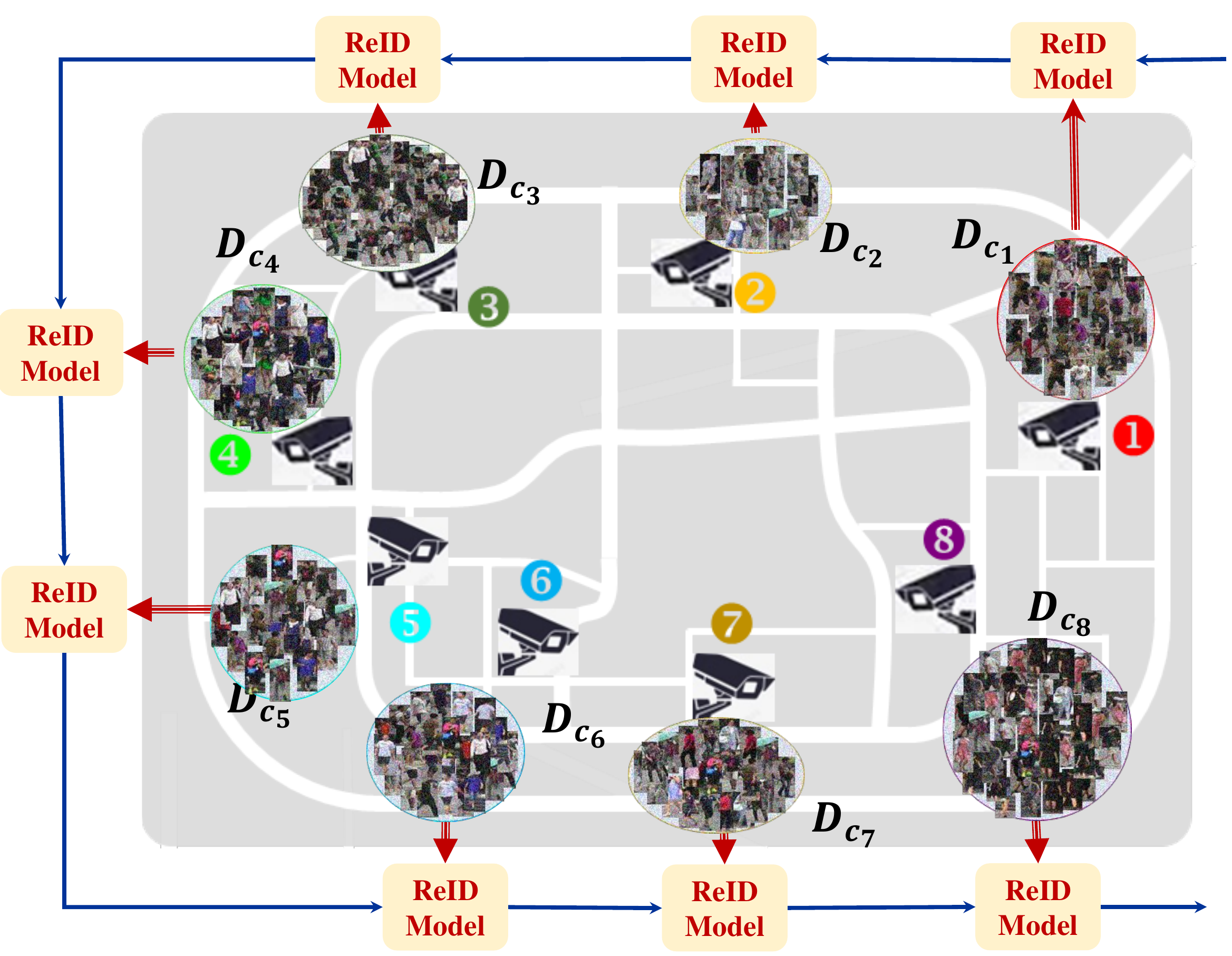}
   \caption{\footnotesize An intuitive description of Camera-Incremental Object Re-Identification (CIOR) with eight camera views, where all camera datasets trained sequentially one after the other.}
   \label{fig:CORI}
\end{figure}

However, using the camera-free stationary dataset to infer a ReID model has the following shortcomings when deploying the ReID system in the real-world.
Firstly, since each camera would generate millions of images every day, merging the images captured from all camera views into a unified dataset needs massive storage, making the ReID model difficult and time-consuming to train.
Secondly, the images captured from each camera view may not be allowed to be stored elsewhere for privacy constraints.
Thirdly, the ReID model inferred on the stationary datasets cannot self-grow and quickly adapt to real-world situations where cameras are dynamically increased.
Finally, it is hard to associate the camera-independent identities, easily generated or annotated, across different camera views to obtain camera-free identities.
The above aspects limit the camera-free ReID mechanism not to be better deployed in the real world,  restricting its generalization and expansion.

In this work, we introduce a novel ReID task named \textbf{C}amera-\textbf{I}ncremental \textbf{O}bject \textbf{R}e-Identification (CIOR)  by continually updating the ReID model with the incoming data of each camera without access to the other cameras.
Unlike the traditional camera-free object ReID, CIOR treats each camera's data separately as a sequential learning problem for different camera views.
An intuitive description of CIOR consisting of eight camera views is shown in Figure~\ref{fig:CORI}, where all the camera datasets are not available during training but encountered sequentially one after the other, and the ReID model trained after each camera view can be used for deployment.
Since CIOR merely considers the identities and images belonging to the current camera and does not access ones from the previous (other) camera views, it thus has two challenges: 1) \emph{Less Discriminative:} as each camera contains a limited number of identities, it is difficult to infer a discriminative ReID model without considering the identities from other camera views; 2) \emph{Catastrophic Forgetting}: without access to the dataset of historical cameras, solely training on the current camera dataset has severe catastrophic forgetting of identities knowledge gained from the historical cameras.

\begin{figure}
  \centering
   \includegraphics[width=0.9\linewidth]{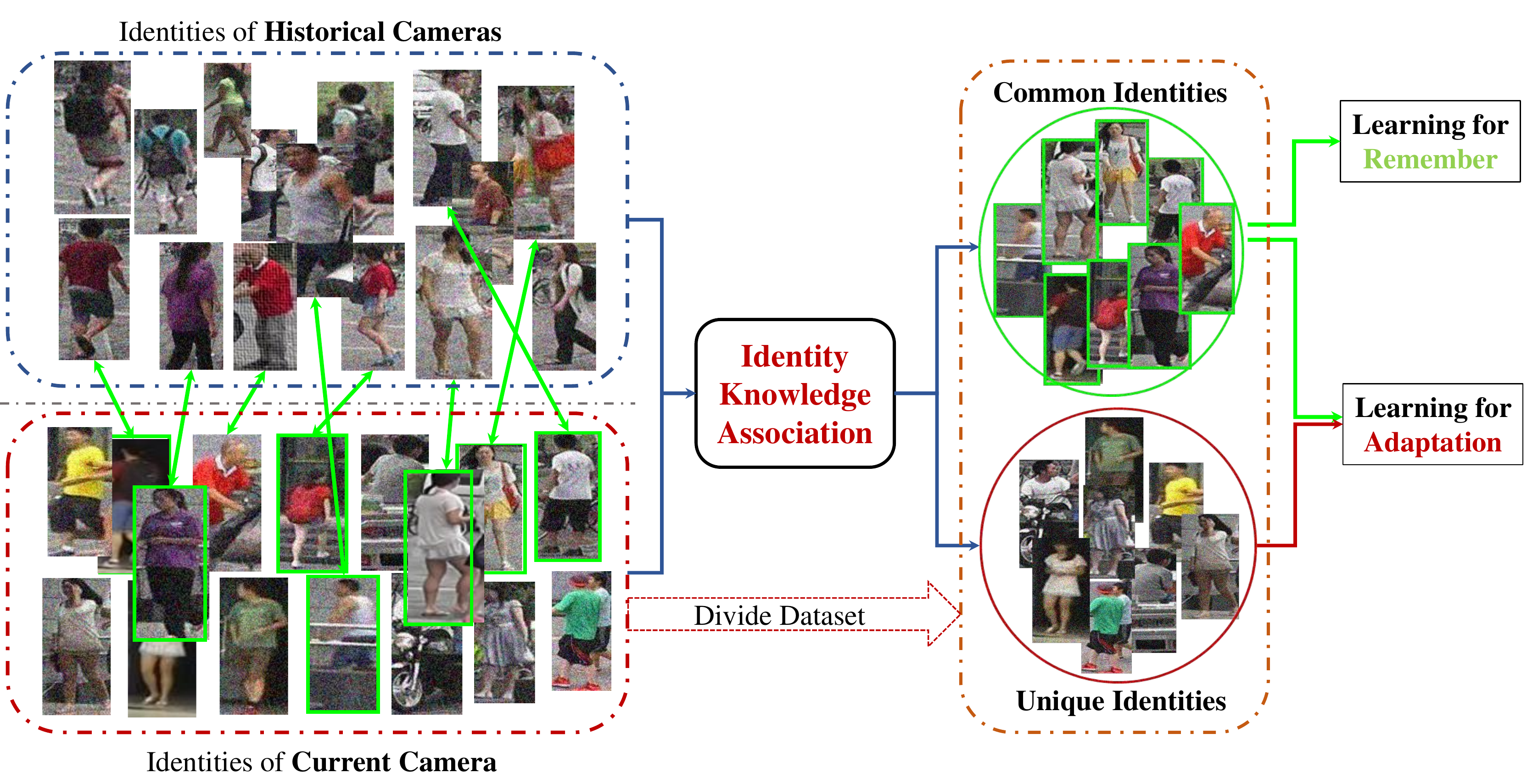}
   \caption{An intuitive description of Identity Knowledge Association for CIOR.Based on the identity knowledge of the historical camera, the identities of current cameras can be divided into two sets: common identities and unique identitis. For incremental learning, the common identities can be applied to remember the historical knolwdge, while the unique identities can be used to infer the newly knowledge.}
   \label{fig:IKA}
\end{figure}

Although some class incremental learning~\cite{KirkpatrickPRVD16,AljundiBERT18,LiH18a,zenke2017continual} and object incremental ReID~\cite{DBLP:conf/aaai/WuG21,DBLP:conf/cvpr/Pu0LBL21,ZhaoTCB021} exist,  CIOR is a more reasonable and challenging task than existing work for real-world object ReID.
Among existing methods, the most related ones to CIOR are class-incremental learning and domain-incremental learning. 
The class-incremental learning assumes that the newly coming images have the disjoint class labels and similar data distribution with the history datasets. 
The domain-incremental learning assumes that each new image has the same class label with different data distribution.
For example, Adaptive Knowledge Accumulation (AKA)~\cite{DBLP:conf/cvpr/Pu0LBL21} defines a domain-incremental person ReID by treating existing ReID benchmarks as an incremental scenario.
By treating the images captured by each camera as domain and the identity as the class, CIOR is the combination of the class-incremental learning and domain-incremental learning, \emph{i.e.,} different camera views contain the different identities (class) with different distributions.
Because the same object might have different identities across different camera views, the different identities under different camera views might represent the same object.
That is to say, different camera views might contain a subset of common identities, as shown in Figure~\ref{fig:IKA}.
In conclusion, the identity of the current camera is implicitly associated with the historical identities instead of being explicitly associated  in traditional class incremental learning, making CIOR to be a more appropriate and challenging setting for object ReID.

Based on the fact that the identities under different camera views might describe the same object, associating and distilling the common identities between the current camera and historical identities would boost the discrimination and benefit to alleviate the catastrophic forgetting.
As shown in Figure~\ref{fig:IKA}, with the identity knowledge association, we can divide the \emph{identities of current camera} as \emph{Common Identities} and \emph{Unique Identities}.
As the \emph{Common Identities} describe the same object from the historical camera views, they can be used to review the historical identity knowledge for alleviating forgetting, and also enhance the discrimination of the ReID model.
Different from the \emph{Common Identities}, the \emph{Unique Identities} is used to adapt the historical ReID model to the current camera for inferring the new knowledge.
Due to the above issue, we propose an Identity Knowledge Evolution (IKE) framework for CIOR, consisting of the Identity Knowledge Association (IKA), Identity Knowledge Distillation (IKD),  and Identity Knowledge Update (IKU).
Identity Knowledge Association(IKA) is proposed to discover the common identities between the current camera and historical camera views with the cyclic-matching strategy.
Then, IKD has applied to distillate historical identity knowledge from common identities and quickly adapt the historical model to the current camera view.
After each camera has been trained, IKU is applied to continually expand the identity knowledge by combining  historical and current identity embeddings.

The major contribution can be summarized as follows:

1) To overcome the shortcoming of existing camera-free object ReID, we introduce a novel ReID task named \textbf{C}amera-\textbf{I}ncremental \textbf{O}bject \textbf{R}e-Identification (CIOR) by continually updating the ReID model merely based on the sequence of camera's dataset;

2) By associating and distilling the common identities between the current camera and historical camera views, we introduce a novel Identity Knowledge Evolution (IKE) framework for CIOR;

3) We adapt the existing Market-1501 and Veri-776 datasets for CIOR, where Market-CL and Veri-CL, consisting of six cameras and fourteen camera views.
The evaluation of two datasets proves the effectiveness of the proposed IKE for CIOR.

\section{Related Work}
\subsection{Object Re-Identification}
Object ReID aims to infer a discriminative ReID model used for retrieving the same objects from the gallery images.
Based on whether using the human annotated labels, object ReID can be divided into supervised object ReID and unsupervised object ReID.
For the supervised object ReID always apply the attention-mechism~\cite{Zhang_2020_CVPR, Chen_2021_ICCV, zhang2021person, xia2019second, chen2019mixed}, part-based description model~\cite{zheng2019pyramidal,fu2019self}, and transform-based description~\cite{He_2021_ICCV,Lai_2021_ICCV} for infer a discrimiantive object description.

As manual annotations are expensive to collect, unsupervised object ReID has attracted much more attention. 
Some researchers use extra labeled images to assist the unsupervised training on unlabeled person ReID by transferring labeled images to the unlabeled domains with GAN-based models~\cite{Chen_2019_ICCV, wei2018person, Liu_2019_CVPR, zhong2018generalizing} or narrowing the distribution gap in feature space~\cite{liu2020domain, Huang2020aaai, dai2021idm}. 
For example, Liu \textit{et al.}~\cite{Liu_2019_CVPR} use three GAN models to reduce the discrepancy between different domains in illumination, resolution, and camera-view, respectively.
To handle the lack of annotation, many methods have been proposed to acquire reliable pseudo labels~\cite{yu2019unsupervised, zeng2020hierarchical,  lin2019bottom, ding2019towards, zheng2021group}. For example, Lin \textit{et al.}~\cite{lin2019bottom} propose a bottom-up unsupervised clustering method that simultaneously considers both diversity and similarity.
Although the above methods can achieve better performance, they all depend on the pre-collected camera-free dataset, limiting the generalization and expansion of the ReID algorithm.
\subsection{Incremental Learning}
Although existing machine learning algorithms have obtained excellent performance for most computer vision tasks, they are all inferred based on the statistic dataset. They have the severe catastrophic forgetting of the historical knowledge when adapting to the new dataset.
To address the above shortcomings, incremental learning or lifelong learning~\cite{delange2021continual} has attracted ever-increasing attention recently.
Based on how and what type of task specific information is used during the sequential training process, existing incremental learning methods can be divided into three classes~\cite{delange2021continual}: Replay methods~\cite{RebuffiKSL17,ShinLKK17,ChaudhryRRE19,Lopez-PazR17}, Regularization-based methods~\cite{KirkpatrickPRVD16,AljundiBERT18,LiH18a}, and Parameter isolation methods~\cite{SerraSMK18,MallyaDL18,MallyaL18}. 
Replay methods~\cite{RebuffiKSL17,ShinLKK17,ChaudhryRRE19,Lopez-PazR17} always store the historical samples after train, which are replayed to train with the current sample to alleviate forgetting.
The shortcoming of replay methods is that they need additional memory space to store the historical samples.
Unlike the replay methods, regularization-based methods~\cite{KirkpatrickPRVD16,AljundiBERT18,LiH18a}  add a regularization term between the weight of the current model and the historical model to consolidate the previous knowledge.
Parameter isolation methods~\cite{SerraSMK18,MallyaDL18,MallyaL18} maintain the independent model parameters for the different tasks to prevent any possible forgetting during the sequential training.

\subsection{Incremental Object Re-identification}
From the type of incremental tasks, incremental learning can be divided into: class-incremental learning~\cite{DBLP:journals/corr/abs-2010-15277}, domain-incremental learning~\cite{DBLP:conf/cvpr/Pu0LBL21}, and task-incremental learning.
The class-incremental learning assumes that the newly coming images have disjoint class labels with the history classes. 
The domain-incremental assumes that the newly coming images have the same class distribution and a large domain gap with the history images.
Inspired by the existing incremental learning methods, a lot of incremental settings are proposed for the object Re-identification~\cite{DBLP:conf/aaai/WuG21,DBLP:conf/cvpr/Pu0LBL21,DBLP:conf/mm/SunM22,DBLP:conf/avss/SugiantoTSCY19}.
For example, Adaptive Knowledge Accumulation (AKA)~\cite{DBLP:conf/cvpr/Pu0LBL21} firstly defines a domain-incremental person ReID by treating existing ReID benchmarks as a incremental scenario, and proposes an Adaptive Knowledge Accumulation by exchange the knowledge between the historical knowledge graph and current knowledge graph.
Using the same incremental setting as AKA, Sun~\emph{et al.}~\cite{DBLP:conf/mm/SunM22} propose a Patch-based Knowledge Distillation to reduce the data distribution discrepancy between the historical and current data.
Furthermore, Wu~\emph{et al.}~\cite{DBLP:conf/aaai/WuG21} design a comprehensive learning objective that accounts for classification coherence, distribution coherence and representation coherence in a unified framework. 

The above-mentioned methods all focus on the domain-incremental person ReID by treating existing ReID benchmarks as different domains for incremental learning.
However, the incremental object re-identification setting based on different datasets has significant difference from the real setting.
A more reasonable incremental setting is to treat the images of each camera as a domain for incremental learning, named Camera-Incremental Object Re-Identification(CIOR).
Specially, by treating the images captured by each camera as domain and the identity as the class, CIOR is the combination of the class-incremental learning and domain-incremental learning, \emph{i.e.,} different cameras contain the different identity (class) and different data distribution, which is more challenge and reasonable than existing incremental object ReID settings.

\section{Camera-Incremental Object Re-identification}
\subsection{Problem Formulation}
We assume that there are $C$ camera views, and the whole dataset is defined as $D=\{D_{1},D_{2},...,D_{C}\}$, where $D_{c}=\{x^{c}_{i},y^{c}_{i}\}(y^{c}_i\in[1,n_c], c\in[1, C])$ denotes the dataset captured from the $c$-th camera view,  $n_{c}$ is the number of identities, $x^{c}_{i}$ and $y^{c}_{i}$ denote the image and its identity label.
Camera-Incremental Object Re-Identification(CIOR) treats each camera's images and identities separately as a sequentially learning problem for different cameras, which leads to a severe catastrophic forgetting for the historical cameras when inferring the current camera.
Take the $c$-th camera view as an example, CIOR aims to infer a robust ReID model $\Phi^{c}$ based on the dataset $D_{c}$, and the historical model $\Phi^{h}$, where $\Phi^{h}$ denotes the historical model inferred from the $(c-1)$-th camera view,
\begin{equation}
\Phi^{c} = \mathcal{A}(\Phi^{h}, D_{c}),
\label{eq:base}
\end{equation}
where $\mathcal{A}(\cdot)$ denotes the ReID algorithm for optimization.

\begin{figure*}
  \centering
   \includegraphics[width=0.8\linewidth]{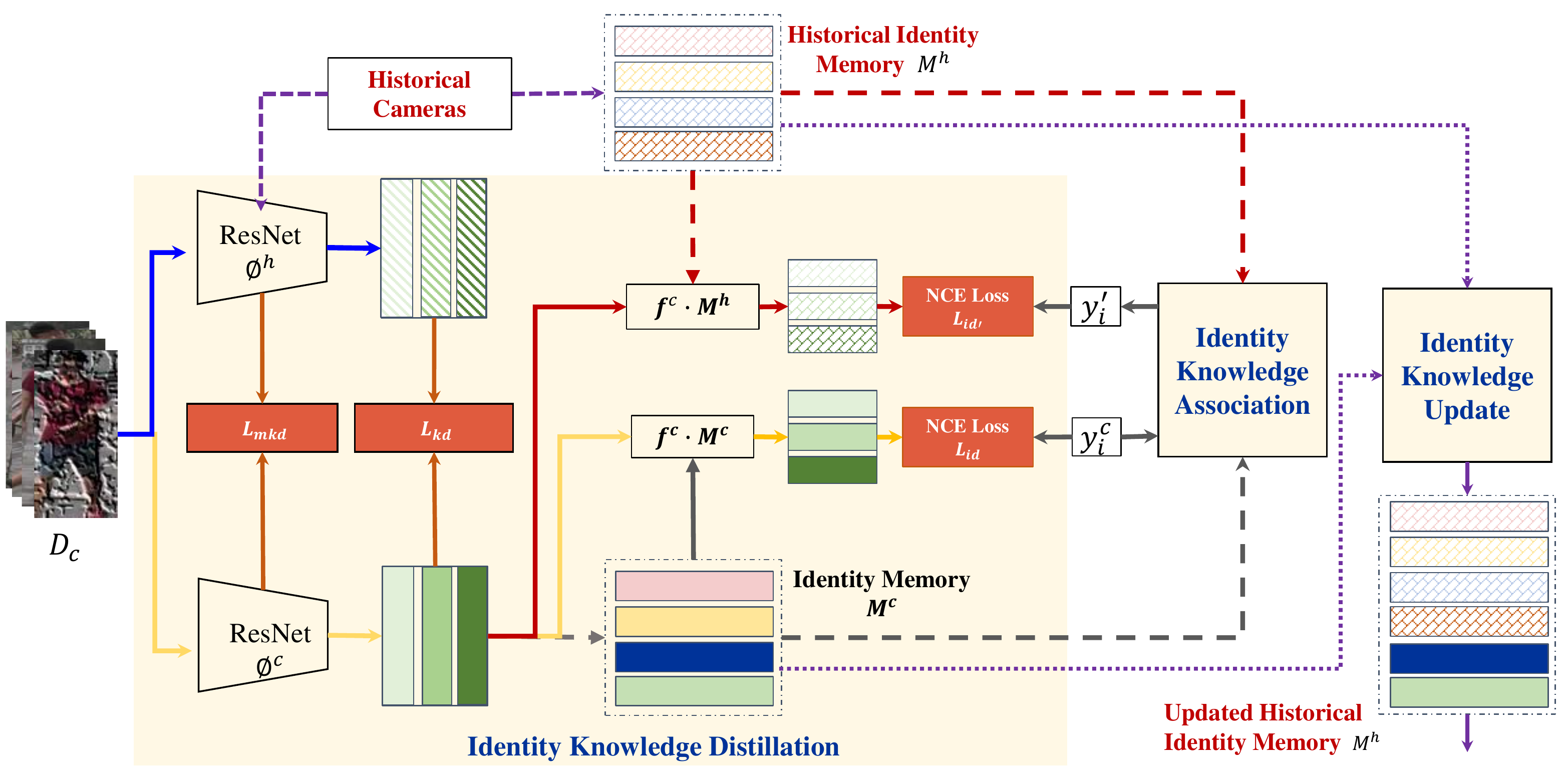}
   \caption{The framework of the Identity Knowledge Evolution framework for the Camera-Incremental Object Re-identification, consisting of the Identity Knowledge Association, Identity Knowledge Distillation, and Identity Knowledge Update.Identity Knowledge Association aims to discover the common identities of the current camera. Identity Knowledge Distillation is used to remember the historical knowledge and infer the new knowledge. Identity Knowledge Update is applied for extend the identity knowledge for incremental learning.}
   \label{fig:IKAE}
\end{figure*}

A simple solution of $\mathcal{A}$ is to treat the historical model $\Phi^{h}$ as an initial model for applying the fine-tune strategy on the dataset $D_{c}$, which is the \emph{Baseline} for CIOR.
As dataset $D_{c}$ cannot describe the identities from other camera views, simply applying to fine-tune would forget the identity knowledge inferred from the previous $(c-1)$ camera views, leading to the inferred model $\Phi^{c}$ a weak descriptive ability.
As the different identity labels under different camera views could describe the same object, associating and distilling the common identities between the current camera and historical identities would boost the discrimination and benefit from alleviating the catastrophic forgetting.
We thus propose a novel Identity Knowledge Evolution framework for CIOR.

\subsection{Identity Knowledge Evolution framework}
After training on the $(c-1)$-th camera view, we can obtain the updated historical identity memory $\mathbf{M}^{h}\in\mathbb{R}^{n_h\times N_d}$, and a historical model $\Phi^{h}$, where $n_h$ is the number of historical identity.
Identity Knowledge Evolution aims to infer a robustness ReID model $\Phi^{c}$ based on the historical identity memory $\mathbf{M}^{h}$, the historical model $\Phi^{h}$, and the dataset $D_{c}=\{x^{c}_{i},y^{c}_{i}\}(y^{c}_i\in[1,n_c])$ of the $c$-th camera view. 
Therefore, Eq.~\eqref{eq:base} can be reformualted as follows:
\begin{equation}
\Phi^{c} = \mathcal{A}(\Phi^{h}, \mathbf{M}^{h}, \mathbf{M}^{c}, D_{c}),
\label{eq:aka}
\end{equation}
where $\mathbf{M}^{c}\in\mathbb{R}^{n_c\times N_d}$ denotes the identity embedding of current $c$-th camera view, which is initialized with the mean feature of each identity.
Specially, given the datasets $D_{c}$, we firstly apply the model $\Phi^{h}$ to extract the feature of each image, and then generate the identity's embedding with the mean of the feature belonging to the same identity. 

Formally, we propose the Identity Knowledge Evolution(IKE) to implement the $\mathcal{A}$ in Eq.~\eqref{eq:aka}.
As shown in Figure~\ref{fig:IKAE}, IKE consists of the backbone module, Identity Knowledge Association(IKA), Identity Knowledge Distillation (IKD), and Identity Knowledge Update(IKU).
IKA is applied for selecting the common identities between the historical identities and identities of the current camera views.
Based on the selected common identities, Identity Knowledge Distillation (IKD) is applied to distillate historical identity knowledge from common identities and quickly adapt the historical model to the current cameras view.
After that, IKU is used to expand historical identity memory by combining current and historical identity memories.
The framework of the Identity Knowledge Evolution is shown in Figure~\ref{fig:IKAE}.

\textbf{Identity Knowledge Association:}
Given the historical identity embedding obtained by the previous camera views, IKA is proposed to select the common identities between the current camera and the historical camera views.
After that, the identities of the current camera can be divided into \emph{Common Identities} and \emph{Unique Identities}.
The \emph{Common Identities} can be used to review the historical identity knowledge for alleviating forgetting, and the \emph{Unique Identities} can be applied to adapt the previous ReID model to the current dataset.
An intuitive motivation of the IKA is shown in Figure~\ref{fig:IKA}.

With the historical identity memory $\mathbf{M}^{h}$ and the current identity memory $\mathbf{M}^{c}$, IKA employs the cycle-consisteny to discover the matching identities between  $\mathbf{M}^{c}$ and $\mathbf{M}^{h}$, \emph{e.g.,} ($y'_{i}, y^{c}_{i}$) denotes an matching pair for the identities $y^{c}_{i}$, where $y'_{i}$ represents the discovered identities of the identities $y^{c}_{i}$ from the historical identity space $\mathbf{M}^{h}$.
For an identity $y^{c}_{i}$, its matched identity $y'_{i}$ is discovered with the cycle-matching between $\mathbf{M}^{c}$ and $\mathbf{M}^{h}$.
Formally,  if the identity embedding $\mathbf{M}^{c}_{y^{c}_{i}}$ has the best matching score with the identity embedding $\mathbf{M}^{h}_{y'_{i}}$, and $\mathbf{M}^{h}_{y'_{i}}$ also has the best matching score with $\mathbf{M}^{c}_{y^{c}_{i}}$,  ($y'_{i}, y^{c}_{i}$) is a matching pair, which is formulated as follows:

\begin{equation}
(y'_{i}, y^{c}_{i})\Longleftrightarrow \left\{
  \begin{aligned}
    \arg\max d(\mathbf{M}^{c}_{y^{c}_{i}};\mathbf{M}^{h})=y'_{i},\\
    \arg\max d(\mathbf{M}^{h}_{y'_{i}};\mathbf{M}^{c})=y^{c}_{i},
   \end{aligned}
   \right.
\label{Eq:m}
\end{equation}
where $d(\mathbf{f},\mathbf{M})$ denotes the cosine distance between the identity embedding $\mathbf{f}$ and the identity memory $\mathbf{M}$, in which a higher score represents a higher similarity.
If not finding the maching identity to $y^{c}_{i}$,  the corresponding identity $y'_{i}=-1$.
After that, two identity labels $y^{c}_{i}$ and $y'_{i}$ can be assigned to the image $x^{c}_i$ for optimization.
For the incremental learning, the identity label $y^{c}_{i}$ is used to adapt the ReID model to the current dataset $D_c$, and  $y'_{i}$ can be applied to reduce the forgetting of the historical knowledge.
With the discovered matching pairs, we can produce a new dataset $D'_{c}=\{x^{c}_{i},y^{c}_{i}, y'_{i}\}$.

\textbf{Identity Knowledge Distillation:}
After obtaining the dataset $D'_{c}=\{x^{c}_{i},y^{c}_{i}, y'_{i}\}$ and the historical model $\Phi^{h}$, IKD is applied to distillate historical identity knowledge with the identity  $y'_{i}$ and quickly adapt the historical model to the current camera view with the identity $y^{c}_{i}$.
Given the image $x^{c}_{i}$, we firstly apply the model $\Phi^{c}$, which is initialized with the historical model $\Phi^{h}$,  to extract its description $\mathbf{f}^{c}_i = \Phi^{c}(x^{c}_i)$.
Next, the cluster contrastive loss based on the identity memory $\mathbf{M}^{c}$ is used for identity classification with Eq.~\eqref{Eq:L},
\begin{equation}
\mathcal{L}_{id}=\sum_{i}^{N'}-\log\frac{\exp(\mathbf{f}^{c}_i\cdot\mathbf{M}^{c}_{y^{c}_i})/\tau}{\sum_{j=1}^{n_c}\exp(\mathbf{f}^{c}_i\cdot\mathbf{M}^{c}_{j})/\tau},
\label{Eq:L}
\end{equation}
where $y^{c}_{i}$ is the corresponding label for the image feature $\mathbf{f}^{c}_{i}$, $\tau$ is a temperature hyper-parameter, and $N'$ is the number of images. 

Inspired by the cluster contrastive learning, the image feature is applied to momentum update the identity memory $\mathbf{M}^{c}$ during backward propagation with Eq.~\eqref{Eq:momentum},
\begin{equation}
\mathbf{M}^{c}_{y^{c}_i}=\omega \mathbf{M}^{c}_{y^{c}_i} + (1-\omega)\cdot \mathbf{f}^{c}_i,
\label{Eq:momentum}
\end{equation}
where $\mathbf{M}^{c}_{y^{c}_i}$ is the $y^{c}_i$-th identity's embedding in identity memory $\mathbf{M}^{c}$, and $\omega$=0.1 is the updating factor.

To reduce the forgetting of the historical identity knowledge during training, the historical identity memory $\mathbf{M}^{h}\in\mathbb{R}^{n_h\times N_d}$ and the identity label $y'_{i}$ are used to optimize the model $\Phi^{c}$ by computing the cluster contrastive loss with the feature $\mathbf{f}^{c}_i$,
\begin{equation}
\mathcal{L}_{id'}=\sum_{i}^{N'}-sgn(y'_i)\log\frac{\exp(\mathbf{f}^{c}_i\cdot\mathbf{M}^{h}_{y'_i})/\tau}{\sum_{j=1}^{n_{h}}\exp(\mathbf{f}^{c}_i\cdot\mathbf{M}^{h}_{j})/\tau},
\label{Eq:L_}
\end{equation}
where $sgn(y)$ is the sign function. $sgn(y)=0$ if $y=-1$. Otherwise $sgn(y)$=1.
As the historical identity memory $\mathbf{M}^{h}$ is generated based on the historical identities, $\mathbf{M}^{h}$ is merely used for computing the loss and not updating.

Furthermore, we apply the knowledge distillation to transfer the  knowledge from the historical model to the current model, which is formulated as follows:
\begin{equation}
\mathcal{L}_{kd}=\sum_{i=0}^{N}sgn(y'_i)||\Phi^{c}(x^c_i)-\Phi^{h}(x^{c}_i)||_2^{2}.
\label{Eq:kd_}
\end{equation}

As Eq.\eqref{Eq:L_} and Eq.~\eqref{Eq:kd_} merely reducing the catastrophic forgetting from the final feature of the model, it has few affect for the middle parameters of the model. 
For example, given an image, although Eq.\eqref{Eq:L_} and Eq.~\eqref{Eq:kd_} can constrain the  historical model and current model to generate the same descriptions,  the generated middle features of those two models would have seriously feature gap. 
To reduce the forgetting of the historical knowledge of middle layers during training, we further construct the knowledge distillation among the middle layers with Eq.\eqref{Eq:mkd},
\begin{equation}
\mathcal{L}_{mkd}=\frac{1}{2}\sum_{i=0}^{N}\sum_{l=2}^{3}sgn(y'_i)||\Phi^{c}_{l}(x^c_i)-\Phi^{h}_{l}(x^{c}_i)||_2^{2},
\label{Eq:mkd}
\end{equation}
where $\Phi^{*}_{l}$ denotes the $l$-th middle features of the model $\Phi^{*}$.
Specially,  $\Phi^{*}_{2}$ and  $\Phi^{*}_{3}$ are the output features of 2-th and 3-th residual blocks of ResNet50, respectively. 

Finally, the total loss is of IKD is:
\begin{equation}
\mathcal{L}_{ikd} = \mathcal{L}_{id} + \mathcal{L}_{id'} + \mathcal{L}_{kd}+\mathcal{L}_{mkd}.
\end{equation}

\textbf{Identity Knowledge Update:}
After training the $c$-th camera view, we can apply the trained model $\Phi^{c}$ to generate its newly identity memory $\mathbf{M}^{c}$ by applying the average of the features belonging to the same identity.
Next, IKU is applied to expand the historical identity memory $\mathbf{M}^{h}$, \emph{i.e.,} it updates the historical identity memory $\mathbf{M}^{h}$ with $\mathbf{M}^{c}$ based on the \emph{updating} and \emph{expansion} rules.

\emph{Updaing rules:} For the identity $y^{c}_j$, if it can find a matched identity $y'_j$ in the history identity memory $\mathbf{M}^{h}$, then updating the $\mathbf{M}^{h}_{y'_j}$ with $\mathbf{M}^{c}_{y^{c}_j}$: 
\begin{equation}
\mathbf{M}^{h}_{y'_j} =  \lambda\times\mathbf{M}^{h}_{y'_j}+(1-\lambda)\times\mathbf{M}^{c}_{y^{c}_j}.
\end{equation}

\emph{Expansion rules:} Otherwise, if it does not discover a matching identity, $\mathbf{M}^{c}_{y^{c}_j}$ is inserted into the historical identity memory $\mathbf{M}^{h}$.
\begin{equation}
\mathbf{M}^{h}=[\mathbf{M}^{h};\mathbf{M}^{c}_{y^{c}_j}],
\end{equation}
where $[;]$ denotes the concatenaton. 
After that, $\mathbf{M}^{h}$ is the updated identity memory used for the next incoming camera dataset. 
Furthermore, the historical model $\Phi^{h}$ is updated with $\Phi^{c}$: $\Phi^{h}$=$\Phi^{c}$.

\section{Experiments}
\subsection{Experimental Setting}
\textbf{Datasets:}
To evaluate the effectiveness of the proposed method, we adopted the existing ReID dataset Market-1501~\cite{zheng2015scalable} and Veri-776~\cite{DBLP:conf/eccv/LiuLMM16} as Market-CL and Veri-CL for Camera-Incremental Object ReID, where Market-CL and Veri-CL consist of six cameras and fourteen cameras, respectively.
Specifically, we adapt the original identity label by independently annotating their identity labels in each camera view.
We discard the cameras whose identities are smaller than 250 for Veri-776.
We feed each camera view dataset sequentially for training, and the standard testing setting is used for evaluation.
The number of identities (\#\emph{ids}) of each camera for Market-CL and Veri-CL are shown in Table~\ref{tab:market} and Table~\ref{tab:veri}, respectively.

\textbf{Implementation Details:} The proposd framework is adopted based on the existing cluster contrastive framework~\cite{dai2021cluster}\footnote{https://github.com/alibaba/cluster-contrast-reid}, which adopts the ResNet-50~\cite{he2016deep} pretrained on ImageNet~\cite{deng2009imagenet} as the backbone.
Inspired by ~\cite{DBLP:conf/cvpr/0004GLL019}, all sub-module layers after layer4-1 are removed, and a GEM pooling followed by batch normalization layer~\cite{DBLP:conf/icml/IoffeS15} and L2-normalization layer is added. 
Therefore, the feature dimension $N_d$ is 2,048.
For the \emph{person ReID}, all input images are resized to 256$\times$128 for training and evaluation with the batchsize of 128.
For the \emph{vehicle ReID}, all input images are resized to 224$\times$224 for training and evaluation with the batchsize of 96.
The temperature coefficient $\tau$ is set to 0.05.
The adam optimizer sets the weight decay as 0.0005, and the learning rate is initially set as 0.00035 and decreased to one-tenth of every 15 epochs up to 30 epochs.

\textbf{Evaluation metrics:}
The metrics $\overline{mAP}$ and $fmAP$ are used to evaluate the performance of the Camera-Incremental Object ReID task.
$\overline{mAP}=\sum_{c=1}^{C}mAP_{c}$ denotes the average of the mean average precision (mAP) obtained by each cameras, where $mAP_{c}$ denotes the mAP after training of the $c$-th camera view, and $C$ is the number of cameras.
$fmAP$ is the mAP after training of all cameras, \emph{i.e.,} $fmAP$=$mAP_{C}$.
Note that the CMC is a popular evaluation metric for the standard Object Re-identification.
However, mAP is a more reasonable and robust evaluation metric than CMC for incremental learning.
We thus report $\overline{mAP}$ and $fmAP$  in this work.

\subsection{Ablation Studies}
This section gives a series of ablation studies to evaluate the influence of the critical components proposed in Identity Knowledge Evolution on the Market-CL datasets.

\begin{table}
\caption{Ablation studies for the Identity Knowledge Evolution.  `*' denotes that the whole identities are used for knowledge distillation in IKD (Eq.~\eqref{Eq:kd_}) and MKD(Eq.~\eqref{Eq:mkd}). }
\label{tab:analysis}
\small
\center
	\begin{tabular}{l|ccccc||cc}
		\toprule
		Methods & $\mathcal{L}_{id}$ &IKA & IKD &MKD &IKU & \emph{fmAP} &$\overline{mAP}$\\ 
		\midrule
		Baseline 	& $\surd$ &  	& &	& 								& 35.2&32.5\\
		IKE-D 		& $\surd$ & $\surd$&$\surd$& & $\surd$			&42.1&37.4\\
		IKE-A 		& $\surd$ &&$\surd$&$\surd$& $\surd$			&44.6&38.7\\
		IKE-U 		& $\surd$ &$\surd$& $\surd$& $\surd$& 			& 47.1&39.4\\
		IKE*      	&$\surd$ &$\surd$ & $\ast$& $\ast$& $\surd$	&47.4&39.5\\
		IKE      	&$\surd$ &$\surd$&  $\surd$& $\surd$& $\surd$	& 48.3&39.8\\
		\bottomrule
        \end{tabular}
\end{table}

\begin{figure}
	\centering
	\includegraphics[width=0.85\linewidth]{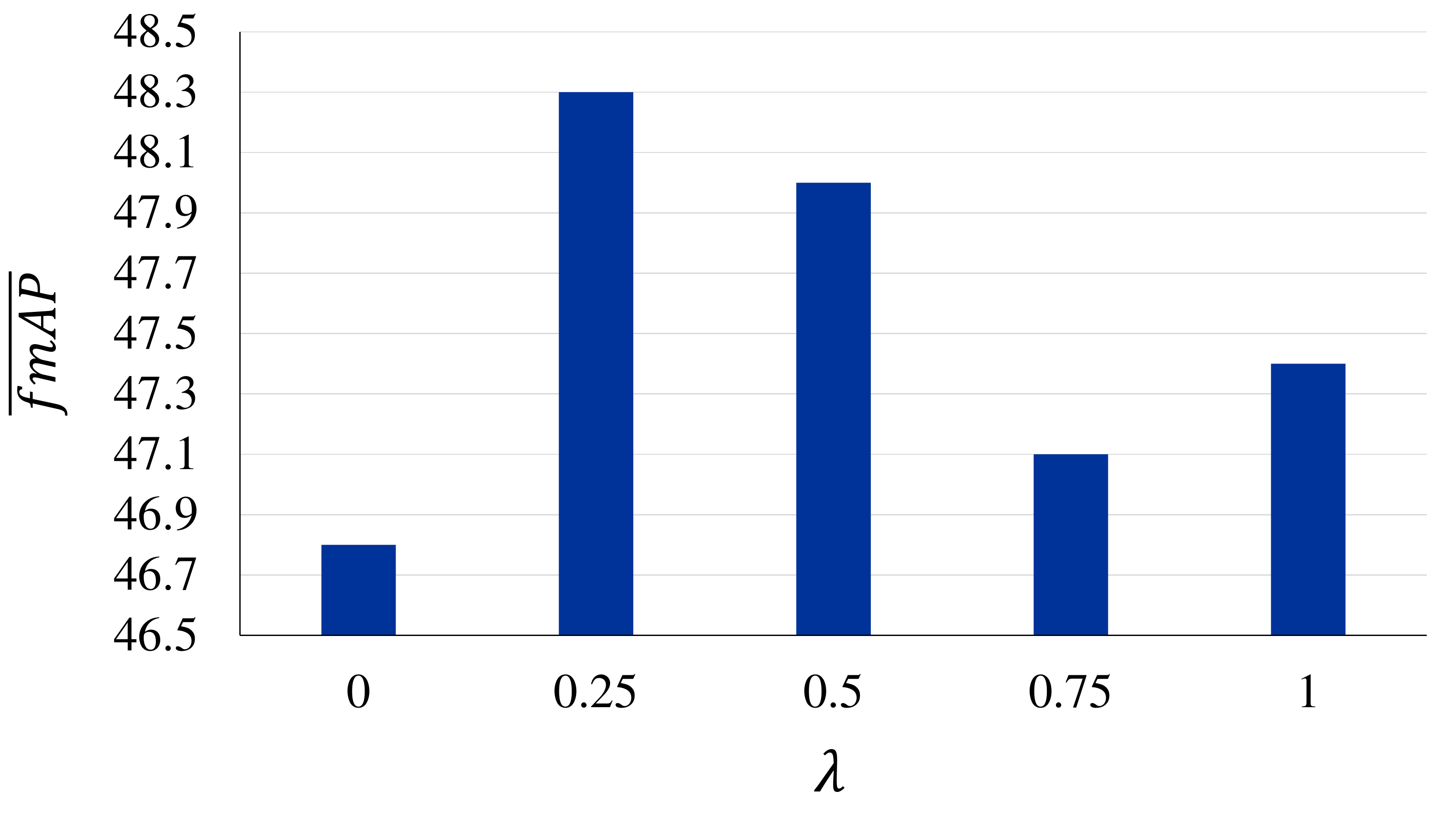}
	\caption{Effect of $\lambda$ for the Identity Knowledge Update..}
	\label{fig:w}	
\end{figure}

\textbf{Effect of  Identity Knowledge Association:}
IKA is proposed to discover the common identities used for identity knowledge distillation. 
To verify the effectiveness of the IKA, we also conduct the comparison by using the whole identities for  the identity knowledge distillation.
For example, `IKE*' uses the whole identities for knowledge distillation in Eq.~\eqref{Eq:kd_} and Eq.~\eqref{Eq:mkd}, and `IKE-A' applies the whole identitis to distillate the historical identity knowledge with Eq.~\eqref{Eq:L_}. 
As shown in Table~\ref{tab:analysis}, `IKE*' and `IKE-A' obtain the \emph{fmAP} of 47.4\% and 44.6\%, which are both lower than 48.3\% obtained by the proposed IKE.
Especially for the `IKE-A', it exists an obvious performance gap with IKE.
The lower performance demonstrates that using the selected common identities can effectively reduce the catastrophic forgetting of the historical identity knowledge.

\textbf{Effect of Multiple Knowledge Distillation:}
Eq.~\eqref{Eq:mkd} is used to transfer the identity knowledge of multiple middle layers from the historical model to the current model.  
As shown in Table~\ref{tab:analysis}, `IKE-D' without considering the constraint $\mathcal{L}_{mkd}$ obtains the \emph{fmAP} of 42.1\%, which is lower than 48.3\% with using the additional constraint $\mathcal{L}_{mkd}$.
The reason is that using the mutiple knowledge distillation can reduce the catastrophic forgetting caused by the middle paramers of each model, which is also complementary to IKD.
Therefore, combining the Multiple Knowledge Distillation and the Identity Knowledge Distillation obtains a superior performance.

\textbf{Effect of Identity Knowledge Update:}
IKU is used to expand the historical identity memory at the end of training of each camera.
We thus make a comparison between the models with/without using IKU, and summarize the related results in Table~\ref{tab:analysis}, where `IKE-U' denotes that IKE uses the identity memory of current camera as the historical identity memory without IKU.
As shown in Table~\ref{tab:analysis}, `IKE-U' obtains the \emph{fmAP} of 47.1\%, which is lower than 48.3\% obtained by the IKE.
The superior performance demonstrates the necessity and importance of IKU for CIOR.

\textbf{Effect of $\lambda$ on IKU:}
For IKU, $\lambda$ is used to merge the historical identity memory $\mathbf{M}^{h}$ and the current identity memory $\mathbf{M}^{c}$.
We thus analyze the effect of $\lambda$  on the Market-CL dataset.
From Figure~\ref{fig:w}, we observe that setting the higher and lower $\lambda$ both obtain a worse performance, \emph{e.g.,} $\lambda$=0.0 and $\lambda$=1.0 obtain the $\overline{mAP}$ of 46.8\% and 47.4\%, respectively.
The worse performance demonstrates that merely considering the limited historical identity knowledge or the current identity knowledge is not suitable for Camera-Incremental Object ReID.
Furthermore, we observe that setting $\lambda$ as 0.25 obtains the best performance, which means that IKU needs to pay more attention to the current identity knowledge most related to the past incoming camera dataset.

\begin{figure}
	\centering
	\includegraphics[width=0.85\linewidth]{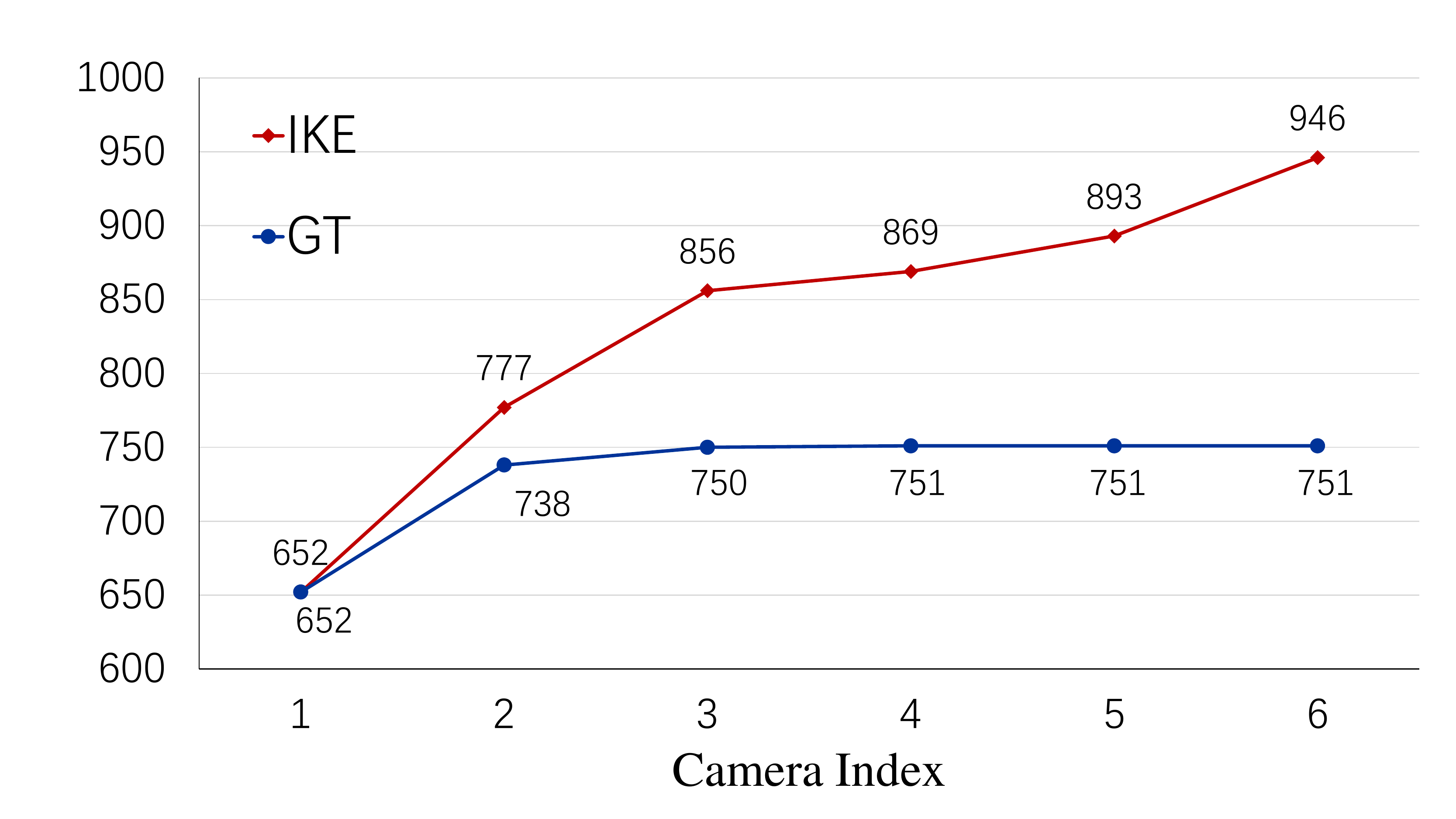}
	\caption{The changing of historical identities' number $D_{h}$ during training on Market-CL.}
	\label{fig:Nh}	
\end{figure}

\begin{figure}
  \centering
   \includegraphics[width=0.8\linewidth]{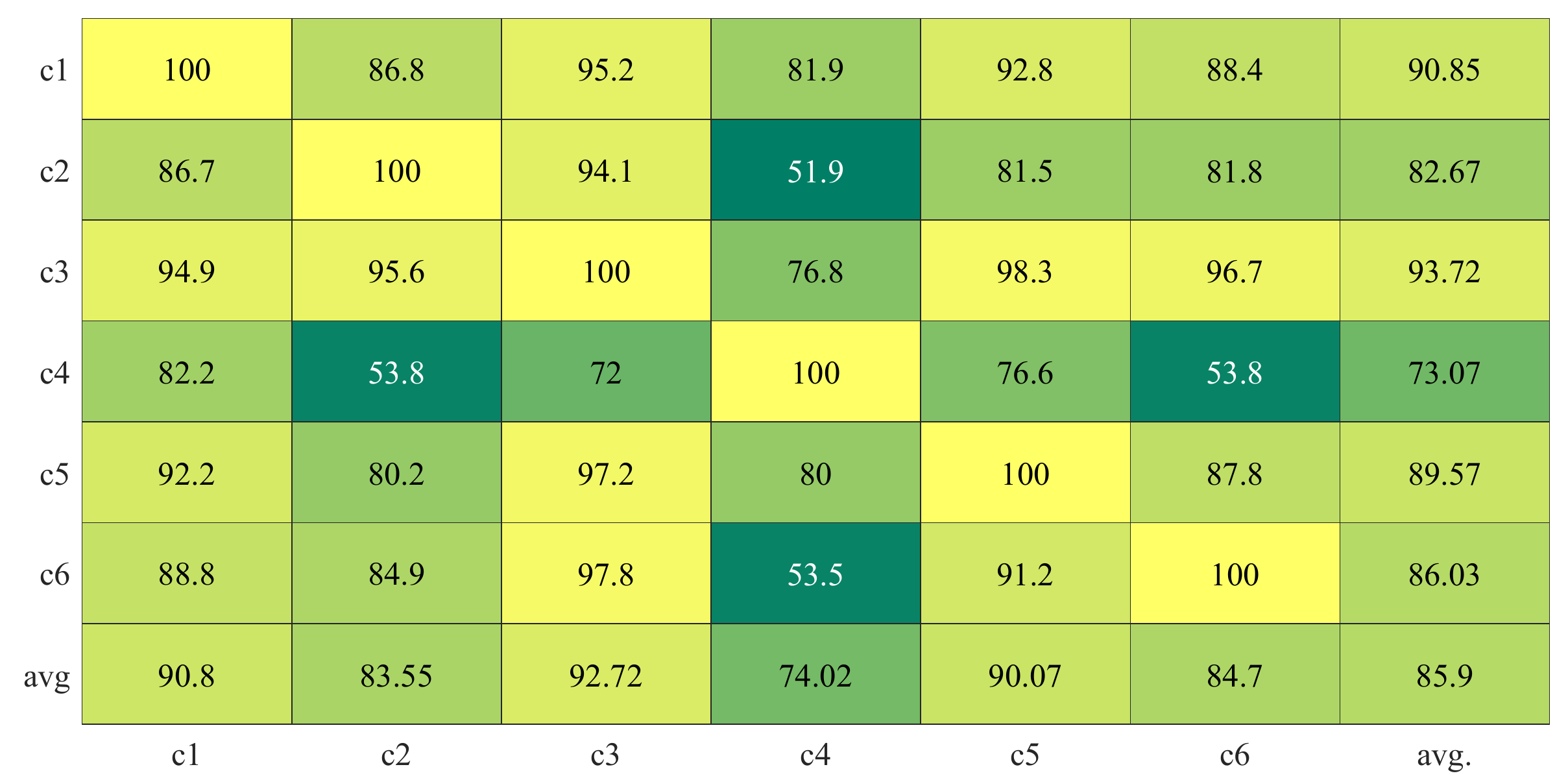}
   \caption{The generated precision matrix $P$ of identity knowledge association between any two camera views(\emph{prec.}(\%)).}
   \label{fig:prec}
\end{figure}

\textbf{Number of historical identities $N_h$:}
IKU merges the historical identity memory $\mathbf{M}^{h}$ and the current identity memory $\mathbf{M}^{c}$ for expanding the identity knowledge, where $\mathbf{M}^{h}\in\mathbb{R}^{N_{h}\times N_{d}}$, and 
$N_{h}$ the number of historical identities after the incremental learning.
 We thus describe the process changing of $N_{h}$ during training, and summarize the related results in Figure~\ref{fig:Nh}.
 `GT' denotes the ground-truth number of identities for the previous $t$-th (camera index) camera views, \emph{e.g.,} 751 is the training identities of the Market-1501.
 From Figure~\ref{fig:Nh}, we observe that the first three cameras can cover the whole identities, \emph{e.g.,} the identities captured by the previous three camera views is 856.
 As shown in Figure~\ref{fig:Nh},  IKU generates the more identities by considering more camera datasets, \emph{e.g.,} IKU generates the final historical identities of 946.

\textbf{Accuracy of Identity Association:}
The critical of Identity Knowledge Evolution is to discover and associate the common identities between current and historical identities with IKA.
We thus analyze the robustness of IKA by computing the ratio of positive matching samples among all discovered matching samples (\emph{prec.}).
Because there are the ground-truth matchings between the first two camera views, we thus compute the precision (\emph{prec.}) for all pair of camera views to generate the precision matrix $P$, where $P[i,j]$ denotes that the ReID model is first trained on the $c_i$-th camera view and then trained on the $c_j$-th camera view to compute the identity association precision on the $c_j$-th camera view.
As shown in Figure~\ref{fig:prec}, IKA obtains a higher precision (\emph{prec.}), \emph{e.g.,} most of the precision is higher than 80\%, which means that the discovered common identities can effectively associate the identities across the historical and current identities space.
The higher precision also demonstrates that the common identities can be used to review identical historical knowledge to alleviate catastrophic forgetting.
From Figure~\ref{fig:prec}, we can observe that the `c4' camera obtains the lowest association precision, \emph{e.g.} treating camera `c4' as the first and second cameras  obtain the average precision of 73.1\% and 74.1\%, respectively. 
The reason is that the camera `c4' contains a few of 241 identities, smaller than other camera views. 

        \begin{table}
		\caption{Comparison of IKE on five different CIOR tasks on Market-CL dataset: $\overline{mAP}$.}
		\label{tab:mike}
        	\centering
        		\centering
            		\begin{tabular}{l|c c c c c || c}
				\toprule
				Tasks & \emph{T1} & \emph{T2} & \emph{T3} & \emph{T4} & \emph{T5} & Avg.\\ 
				\midrule
				Baselines&32.3&31.7 &32.7&33.3&32.3&32.5\\ 
				KD~\cite{HintonVD15} 	&35.6	& 36.2 & 36.5 & 32.3 & 38.7 & 35.9 \\
				CRL~\cite{ZhaoTCB021} 	& 32.7 & 32.6 & 33.1 & 30.7 & 32.3 & 32.5 \\
				LWF~\cite{LiH18a} 	& 35.3 & 35.7 & 35.8 & 33.5 & 36.9 & 35.4 \\
				EWC~\cite{KirkpatrickPRVD16}  	& 35.7 & \textbf{39.8} & \textbf{40.7} & 33.8 & 42.4 & 38.5\\
				MAS~\cite{AljundiBERT18}   	& 35.7	& 38.8 & 38.8 & 34.1 & 42.0 & 37.9\\
				\midrule
				IKE     	& \textbf{39.8} & 39.3 & 39.6 & \textbf{34.3} & \textbf{44.7} & \textbf{39.5}\\
				\bottomrule
            		\end{tabular}
	\end{table}

        \begin{table}
		\caption{Comparison of IKE on five different CIOR tasks on Market-CL dataset: \emph{fmAP}.}
		\label{tab:mike}
        	\centering
            		\begin{tabular}{l|c c c c c || c}
				\toprule
				Tasks & \emph{T1} & \emph{T2} & \emph{T3} & \emph{T4} & \emph{T5} & Avg.\\ 
				\midrule
				Baselines& 43.2 & 41.9  & 42.7 & 39.6 & 46.4 & 42.7\\ 
				KD~\cite{HintonVD15} 	& 40.8	& 39.6 & 38.3 & 40.9 & 39.8 & 39.9 \\
				CRL~\cite{ZhaoTCB021} 	& 35.8 & 38.2 & 32 & 37.4 & 34.1 & 35.5 \\
				LWF~\cite{LiH18a} 	& 48.9 & 39.6 & 36.4 & 42.3 & 38.1 & 41.1 \\
				EWC~\cite{KirkpatrickPRVD16}  	& 41.5 & \textbf{45.4} & \textbf{47.3} & 44.1 & 48.8 & 45.4\\
				MAS~\cite{AljundiBERT18}   	& 44.9	& 44.3 & 42.9 & 44.3 & 45.1 & 44.3\\
				\midrule
				IKE     	& \textbf{48.3} & 44.8 & 47.2 & \textbf{44.4} & \textbf{51.8} & \textbf{47.3}\\
				\bottomrule
            		\end{tabular}
	\end{table}

\textbf{Effect of the mulitiple cameras's orders:}
For CIOR, the cameras' dataset always consists of several cameras, \emph{e.g,} Market-CL and Veri-CL contains six and fourteen camera views, respectively.
As each camera view consists of the different identities and images, the order of cameras may affect the performance of CIOR.
To further evaluate whether the proposed methods are insensitive to the camera's order, we randomly generate five different CIOR settings for Market-CL with the different orders of cameras: 

1) \emph{Task1}: c1$\rightarrow$c2$\rightarrow$c3$ \rightarrow$c4$\rightarrow$c5$\rightarrow$c6;

2) \emph{Task2}: c1$\rightarrow$c6$\rightarrow$c5$\rightarrow$c2$\rightarrow$c4$\rightarrow$c3;

3) \emph{Task3}: c6$\rightarrow $c3$ \rightarrow$c4$\rightarrow$c5$\rightarrow$c1$\rightarrow$c2;

4) \emph{Task4}: c4$\rightarrow$c2$\rightarrow$c6$\rightarrow$c5$\rightarrow$c3$\rightarrow$c1;

5) \emph{Task5}: c3$\rightarrow$c1$\rightarrow$c4$\rightarrow$c5$\rightarrow$c2$\rightarrow$c6;  \\
where \emph{Task1} is the standard order of cameras in the original dataset.
From Table~\ref{tab:mike}, we observe that different camera orders affect the performance.
For example, IKE, EWC, and MAS all obtains higher performance on \emph{Task5} than the rest four tasks, indicating that the order of the incoming camera's dataset having obvious effect on CIOR. 
Compared with the existing methods, the proposed IKE obtains the best performance in three of the five tasks, \emph{e.g.,} \emph{Task1}, \emph{Task4}, and \emph{Task5}.
However, the proposed IKE obtains the average performance of 39.5\%/47.3\% for $\overline{mAP}$/\emph{fmAP}, which is higher than the second best performance of 38.5\%/ 45.4\% obtained by the EWC.

\begin{figure}
   \includegraphics[width=0.9\linewidth]{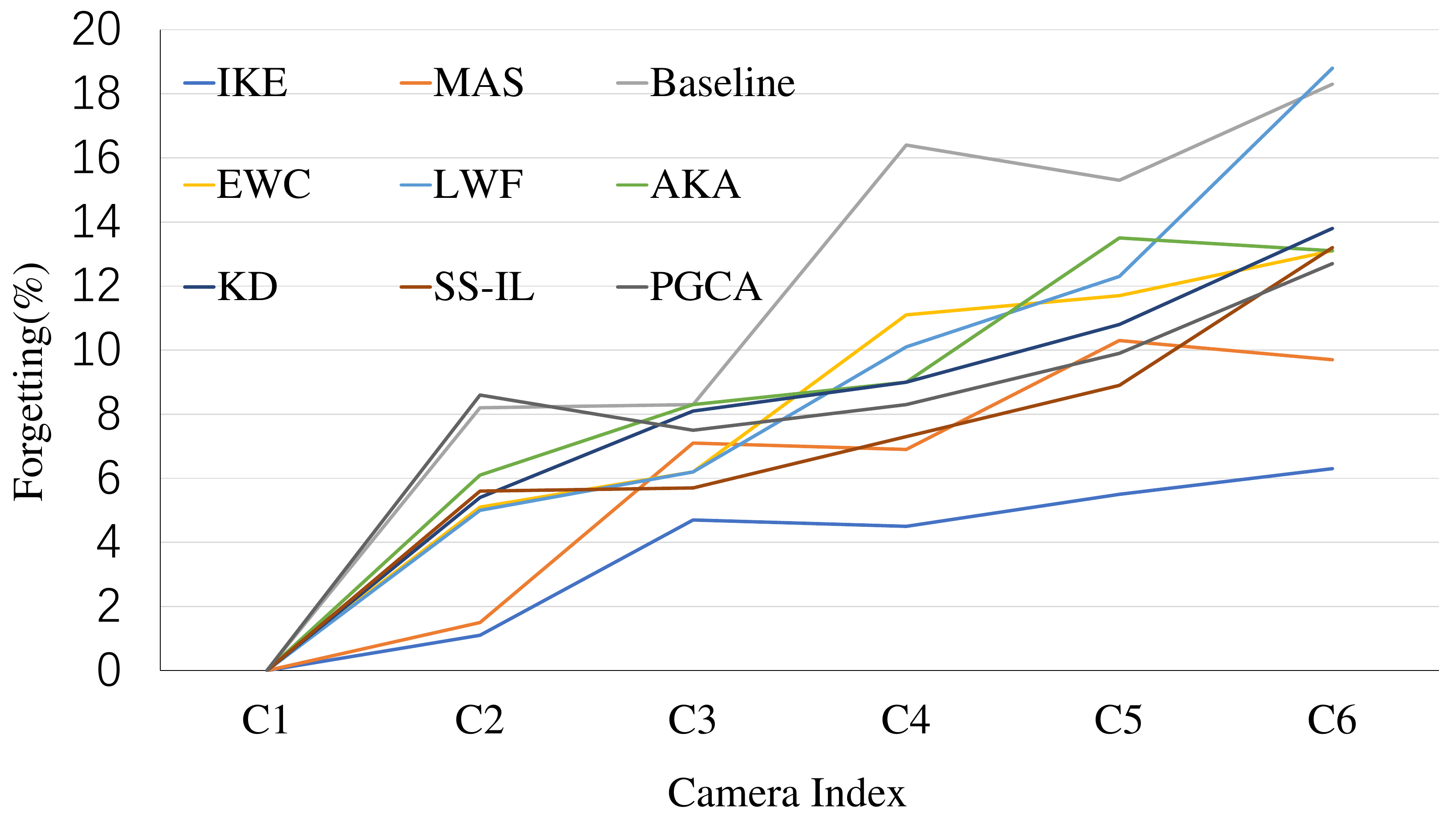}
   \caption{Forgetting trend on Market-CL. The lower forgetting ratio, the better.}
   \label{fig:forget} 
\end{figure}

\begin{table*}
\caption{Comparion with existing methods on Market-1501 datasets. `CI' and `DI' denote `Class-Incremental' and `Domain-Incremental', respectively. `M=2' denotes that storing two images of each historical identity.}
\label{tab:market}
\centering
\begin{tabular}{c|c|cccccc|cc}
\toprule
Settings                                                                                     & Methods & C1 & C2 & C3 & C4 & C5 & C6&\emph{fmAP}& $\overline{mAP}$ \\
\midrule
&\#ids & 652 & 541 & 694 & 241 & 576 & 558 & -&-\\
\midrule
-&Baseline~\cite{dai2021cluster}& 29.7 & 27.2 & 37.6 & 30.0 & 33.5 & 36.3 & 36.3&32.4\\
-&KD~\cite{HintonVD15} &29.7& 30& 37.8&37.4 & 38& 40.8& 40.8& 35.6\\
\midrule
\multicolumn{1}{c|}{\multirow{4}{*}{CI}}                                          & EWC~\cite{KirkpatrickPRVD16}     & 29.7&30.3 & 39.7&  35.3&37.1&41.5& 41.5&35.7\\
\multicolumn{1}{c|}{}                                                                            & MAS~\cite{AljundiBERT18}     &  29.7 &33.9 &38.8 &39.5 &38.5 & 44.9 &44.9&37.5            \\
\multicolumn{1}{c|}{}                                                                            & LWF~\cite{DBLP:conf/eccv/LiH16}     & 29.7&30.4 &39.7 &36.3 &36.5 &38.8 & 38.8&35.3           \\
\multicolumn{1}{c|}{}                                                                            & CRL~\cite{ZhaoTCB021} &29.7& 28.3& 38& 30.5& 33.6& 35.8&35.8 &32.65\\
\multicolumn{1}{c|}{}                                                                            & SS-IL~\cite{AhnKLBKM21}   & 29.7&29.8 & 40.2&39.1 & 39.9 &41.4& 41.4 &35.0\\
\midrule
DI                                                                            & AKA~\cite{DBLP:conf/cvpr/Pu0LBL21}     & 25.5&29.3 &36.2 &37.4 &35.3 &41.5&41.5 & 34.2\\
\midrule
Exempler                                                                                  & CRL(M=2)~\cite{ZhaoTCB021} & 29.7 & 32.3& 39.3& 39.7& 40.2& 41.6&41.6&37.1 \\
\midrule
\multirow{2}{*}{\begin{tabular}[c]{@{}l@{}}CI+ DI\end{tabular}} & PGCA ~\cite{LinZQNGLT22}  & 29.7&26.8&38.4 &38.1 & 38.9& 41.9& 41.9&35.6\\
                                                                                                & \textbf{IKE}     &  29.7& \textbf{34.3}&\textbf{41.2} &\textbf{41.9} & \textbf{43.3}& \textbf{48.3}&\textbf{48.3}&\textbf{39.8}\\
\midrule
-&UpBound & 29.7  & 35.4 &45.9&46.4&48.8&54.6&54.6&43.15\\
\bottomrule
\end{tabular}
\end{table*}

\begin{table*}
\caption{Comparison on Veri-CL datasets.}
\label{tab:veri}
\centering
\begin{tabular}{l|cccccccccccccc||cc}
\toprule
Methods  & c1   & c2   & c3   & c4   & c5   & c6   & c7   & c8   & c9   & c10  & c11  & c12  & c13  & c14  &\emph{fmAP} & $\overline{mAP}$\\
\midrule
 \#ids        & 316  & 273  & 296  & 294  & 268  & 267  & 267  & 313  & 314  & 465  & 356  & 350  & 340  & 335  &      &           \\
\midrule
Baseline & 16.7 & 21.7 & 26.3 & 27.2 & 30.2 & 28.5 & 27.3 & 26.7 & 25.7 & 27.4 & 27.4 & 27.8 & 29   & 24.2 & 24.2 & 26.1     \\
\midrule
AKA~\cite{DBLP:conf/cvpr/Pu0LBL21}	& 14.12 & 21.4 & 26.4 & 25.3 & 28.1 & 28.1 & 26.3 & 23.7 & 24.9 & 27.7 & 26.7 & 26.7 & 28.8   & 24.1 & 24.1 &  25.1\\ 
CRL~\cite{ZhaoTCB021}      & 16.7 & 21.2 & 26.9 & 26.5 & 29.8 & 26.9 & 28.5 & 25.8 & 27.4 & 26.9 & 27.3 & 27.6 & 30.4 & 22.8 & 22.8 & 26.0     \\
KD~\cite{HintonVD15}       & 16.7 & 22.3 & 26.6 & 28.7 & 30.4 & 31.1 & 31   & 30.8 & 31   & 31.1 & 31.6 & 31.9 & 32.8 & 29.1 & 29.1 & 28.9  \\
LWF~\cite{DBLP:conf/eccv/LiH16}      & 16.7 & 22.4 & 28.1 & 30.2 & 32.2 & 30.6 & 29.4 & 28.5 & 28.5 & 29.6 & 29.1 & 29.7 & 31.4 & 27.5 & 27.5 & 28.1  \\
EWC~\cite{KirkpatrickPRVD16}      & 16.7 & 21.7 &29.7 & \textbf{30.7} & \textbf{33.3} & \textbf{34.2} & 30.8 & 31.2 & 30.3 & 31.3 & 32.6 & 33.1 & 33.1 & 25   & 25   & 29.6     \\
MAS~\cite{AljundiBERT18}      & 16.7 & 22.6 & 27.8 & 30.2 & 31.7 & 32.9 & 30.5 & 30.9 & 30.4 & 32.4 & 32.9 & 32.7 & 32.6 & 26.3 & 26.3 & 29.3  \\
SS-IL~\cite{AhnKLBKM21}     & 16.7&\textbf{24.6}&27.5&29.8&30.4& 29.1&28.6&27.6&28.4&27.8&27.7& 26.8&28.6&25.3&25.3 &27.1 \\
PGCA~\cite{LinZQNGLT22}      & 16.7&24.2 &\textbf{32.1}&30.4&33&29.4&31.4&27.5&30.2&30.5&29.6&33.2&31.8&25.1& 25.1&28.9 \\
\midrule
IKE      & 16.7 & 23.6 & 26.3 & 28.5 & 31.1 & 31.8 & \textbf{32.5} & \textbf{32.9} & \textbf{32.6} & \textbf{33.2} & \textbf{34.5} & \textbf{35}   & \textbf{35.3} & \textbf{33}   & \textbf{33}   & \textbf{30.5}  \\  
\bottomrule
\end{tabular}
\end{table*}

\textbf{Forgetting trend:}
The criticial of the incremental learning is to reduce the forgetting ratio of existing object ReID methods.
We thus analyze the forgetting ratio of the compared methods, where the \emph{Forgetting} is the performance gap between each method and the upbound performance.
As shown in Figure.~\ref{fig:forget}, the proposed IKE method achieves the lowest forgetting ratio during the incremental learning.

\subsection{Comparison with existing methods}
In this section, we compare the proposed IKE and the existing incremental learning methods on Market-CL and Veri-CL datasets, and summarize the related results in Table~\ref{tab:market}, and Table~\ref{tab:veri}.
Existing compared methods can be divided  into: class-incremental(CL) methods( CRL~\cite{ZhaoTCB021}, LWF~\cite{LiH18a},EWC~\cite{KirkpatrickPRVD16}, MAS~\cite{AljundiBERT18}, and SS-IL~\cite{AhnKLBKM21}), domain-incremental(DL) methods(AKA~\cite{DBLP:conf/cvpr/Pu0LBL21}), exempler-based method(CRL~\cite{ZhaoTCB021}), and domain- and class- incremental methods(PGCA~\cite{LinZQNGLT22}).

As shown in Table~\ref{tab:market},  `Baseline' model obtains the \emph{fMAP} of 36.3\% by merely conducting the fine-tune on each camera, having a large gap with the upbound of 54.6\%.
The large gap demonstrates that there exists a severe catastrophic forgetting by simply fine-tuning each camera dataset.
Compared with the `Baseline', the proposed IKE obtains large improvement on two type of evaluation metrics, \emph{e.g.,} the $fmAP$ and $\overline{mAP}$ are improved from the 36.3\%/32.4\% to 48.3\%/39.8\% for Market-CL dataset.

Among the class-incremental methods, MAS~\cite{AljundiBERT18} obtains a best performance, \emph{e.g.,} 44.9\%/37.5\% for \emph{fmAP}/$\overline{mAP}$.
Compared with MAS, IKE obtains an improvement of 3.4\%/2.3\% for \emph{fmAP}/$\overline{mAP}$.
We also observed that the domain-incremental method AKA obtains a similar performance with EWC, lower than MAS and the proposed IKE.
Among all existing methods, PGCA is the most related work to ours, which is proposed for Class-Incremetal Unsupervised Domain Adaptation, consisting of domain-incremental and class-incremental.
PGCA discovers the novel class by computing the cumulative probability (CP) of target samples regarding source classes.
Different from PGCA, IKE applies an Identity Knowledge Association to discover the the common identities apart from the new identitis.
As shown in Table~\ref{tab:market}, the proposed IKE obtains a significant improvement upon the PGCA, \emph{e.g.,} improving \emph{fmAP}/$\overline{mAP}$ from 41.9\%/35.6\% to 48.3\%/39.8\%.
The superior performance demonstrates the effectiveness of the proposed IKE. 

As shown in Table~\ref{tab:veri}, the proposed IKE still obtains the superior performance than existing methods on Veri-CL datasets.
The superior performance demonstrates the effectiveness of the proposed IKE for Camera-Incremental Object ReID.

\section{Conclusion}
In this work, we introduce a novel ReID task named Camera-Incremental Object Re-identification (CIOR) by continually updating the ReID model with the incoming data of the current camera.
Unlike the traditional camera-free object ReID, CIOR treats each camera's data separately as a sequentially learning problem for different cameras, leading to severe catastrophic forgetting for the historical cameras when inferring the current cameras.
Furthermore, we propose a novel Identity Knowledge Evolution to associate and distillate the historical identity knowledge for alleviating forgetting.
The evaluation of two adapted benchmarks, Market-CL and Veri-CL, validated the effectiveness of the IKE for CIOR.

Although the proposed IKE is an effective method for CIOR, it performs slightly worse than the regularization-based methods on some causes, \emph{e.g.,} EWC obtains higher performance than IKE on \emph{Task2} and \emph{Task3}.
As the IKE can be treated as the knowledge distillation-based methods, it might be complement to the regularization-based methods.
In the future, we will try to combine the advantages of these two types of algorithms to propose a more robust algorithm for CIOR.

\newpage


\newpage
\bibliographystyle{ACM-Reference-Format}
\bibliography{egbib}


\end{document}